\documentclass[letterpaper, 10 pt, conference]{ieeeconf}  
\IEEEoverridecommandlockouts                              
\usepackage[absolute]{textpos}
\newcommand{\copyrightstatement}{
    \begin{textblock}{13.17}(1.36, 15.22)    
         \noindent
         \footnotesize
         \copyright  © 2019 IEEE.  Personal use of this material is permitted.  Permission from IEEE must be obtained for all other uses, in any current or future media, including reprinting/republishing this material for advertising or promotional purposes, creating new collective works, for resale or redistribution to servers or lists, or reuse of any copyrighted component of this work in other works.
    \end{textblock}
}

\usepackage{graphics} 

\title{\LARGE \bf
Decay Replay Mining to Predict Next Process Events
}

\usepackage{cite}
\usepackage{makecell}
\usepackage{amsmath,amssymb,amsfonts}
\usepackage{algorithmic}
\usepackage{graphicx}
\usepackage{textcomp}
\usepackage{multirow}
\let\labelindent\relax
\usepackage{enumitem}

\usepackage{floatrow}
\DeclareFloatFont{tiny}{\tiny}
\usepackage[table,xcdraw, pdftex]{xcolor}

\def\BibTeX{{\rm B\kern-.05em{\sc i\kern-.025em b}\kern-.08em
    T\kern-.1667em\lower.7ex\hbox{E}\kern-.125emX}}

\usepackage{caption}
\usepackage[skip=2pt,font=footnotesize]{caption}

\author{{Julian~Theis,~\IEEEmembership{Graduate Student Member,~IEEE}
and~Houshang~Darabi,~\IEEEmembership{Senior Member,~IEEE}}
\thanks{J. Theis and H. Darabi are with University of Illinois at Chicago, Department of Mechanical and Industrial Engineering, 842 West Taylor Street, Chicago, IL 60607, United States. H. Darabi is the corresponding author. E-mail: \{jtheis3, hdarabi\}@uic.edu.}}

\begin{document}
\copyrightstatement
\maketitle
\thispagestyle{empty}
\pagestyle{plain}
\begin{abstract}
In complex processes, various events can happen in different sequences. The prediction of the next event given an a-priori process state is of importance in such processes. 
Recent methods have proposed deep learning techniques such as recurrent neural networks, developed on raw event logs, to predict the next event from a process state. However, such deep learning models by themselves lack a clear representation of the process states. At the same time, recent methods have neglected the time feature of event  instances. 
In this paper, we take advantage of Petri nets as a powerful tool in modeling complex process behaviors considering time as an elemental variable. 
We propose an approach which starts from a Petri net process model constructed by a process mining algorithm. We enhance the Petri net model with time decay functions to create continuous process state samples.
Finally, we use these samples in combination with discrete token movement counters and Petri net markings to train a deep learning model that predicts the next event.
We demonstrate significant performance improvements and outperform the state-of-the-art methods on nine real-world benchmark event logs.
\end{abstract}
\begin{keywords}
Business Process Intelligence,
Decay Functions,
Deep Learning,
Petri Nets,
Neural Networks,
Operational Runtime Support,
Predictive Process Management,
Process Mining
\end{keywords}

\section{Introduction}
With the ongoing development of digitizing and automatizing industries along with the steady increment of interconnected devices, we can project more interactions onto processes \cite{intro_iot_bpi, intro_iot_bpi2}. These processes can represent procedures in different industries such as retail \cite{retail}, software development \cite{software}, healthcare \cite{healthcare}, network management \cite{networkmanagement}, project management \cite{darabirev2}, or manufacturing \cite{manufacturing}. One illustrative example is the process of a customer loan application in financial institutes \cite{data_bpic12}. An applicant can request money for specific purposes. The application then undergoes several process steps  such as \textit{negotiation}, \textit{request validation}, \textit{fraud assessment}, \textit{offer creation} and/or \textit{application rejection}. Each step  of the process utilizes different institutional resources such as employees, customer records, IT systems, or third-party resources to check the creditworthiness of applicants. Though trivial, the process gets complex with an increasing number of applications and requirements of the institute.

While traditional process mining is primarily concerned with the discovery, analysis, and monitoring of processes, predictive process management gains momentum by enhancing process models. Predictive process management plays an important role in the areas mentioned earlier. Knowing when specific situations occur, or in which state a process will be next, is important to meet qualitative and/or quantitative requirements of businesses and organizations.

Many businesses deploy Process-Aware Information Systems such as workflow management systems, case-handling systems, enterprise information systems, enterprise resource planning, and customer relationship management systems. These are software tools which manage and execute operational processes involving people, applications, and/or information sources based on process models \cite{intro_pais}. Such systems record events associated to different process steps along with time and other related information which can be utilized for predictive process management.  Typical use cases comprise the prediction of the next event, forecasting of a process' final state, or time interval prediction of future events \cite{intro_predictiveprocessmanagement}. Predicting the next event elicits special attention since it gives organizations the ability to forecast process deviations. This type of early detection is essential for intervenability before a process enters risky states \cite{litrev_breuker}. Moreover, predictive process management assists businesses in resource planning and allocation, providing insights on the condition of a process to fulfill for instance service-level agreements \cite{intro_applications, leitner2009runtime, litrev_tax}. 

With this motivation, a range of different methods have been proposed on predicting the events in process sequences. Most recent advances are made in utilizing different deep learning architectures such as Long Short-Term Memory (LSTM) neural networks and stacked autoencoders \cite{litrev_evermann, litrev_tax, litrev_khan, litrev_mehdiyev}.
However, these techniques do not discover process models at first, but perform their predictions on the raw event logs. 
This makes decision making hard to understand and difficult to explain, which is crucial to discover the weaknesses of a process.
Furthermore, since neural networks are not infallible \cite{intro_infallible}, commonsense knowledge and obvious logical policies are suggested to be introduced into a deep learning model from the beginning to reduce potential vulnerability. This knowledge is easy to obtain from process discovery algorithms. Therefore, modeling processes from scratch using neural networks is costly and partially redundant. 
Thus, one of the research questions is how to retain process models like Petri nets (PN) \cite{PNBasic} with its logic, interpretability, and comprehensibility \cite{processmining, interpretability, interpretability2, interpretability3}, and combine it with the strengths of deep learning towards more interpretable models to improve performance at the same time. 

A further research motivation arises due to weaknesses of recent predictive methods. Some of the state-of-the-art algorithms do not consider event timestamps as features at all \cite{litrev_breuker, litrev_evermann}. However, the duration between two events and/or sequences of events might be correlated with a future process outcome. Therefore, we suggest taking event times into account for predictive modeling.

In the current work, we propose an innovative method to predict the  next event  of a running process case which engages with the issues mentioned above. 
We first leverage a state-of-the-art process mining algorithm to discover a PN based process model from an event log. Then, we enhance the process model with time decay functions. In this way, we can create continuous and timed state samples which we finally couple with process resources to train a neural network for the prediction of the  next event. 
We call this approach \textit{Decay Replay Mining - Next  TrAnsition  Prediction} (DREAM-NAP).
By taking this approach, we demonstrate significant performance improvements. Our method outperforms the state-of-the-art techniques on all of the popular benchmark datasets. 

This paper is structured as follows. Section \ref{sec:relatedwork} discusses related work and most recent advances in the next event prediction of business processes. We introduce preliminaries in Section \ref{sec:prelim}. Section \ref{sec:approach} focuses on the proposed approach, especially on the decay function modeling in PNs and the deep learning architecture. Section \ref{sec:evaluation} evaluates the approach against different existing methods. Finally, we conclude the paper and discuss future work in Section \ref{sec:conclusion}.

\section{Related Work}\label{sec:relatedwork}
The application of deep learning on predictive business process mining has grown enormously during recent years. Researchers have shown the applicability of machine and deep learning on several target variables such as the remaining time of running cases \cite{relatedwork_example_remainingtime}, forecasting time of events \cite{intro_navarin}, and predicting upcoming events in running processes while utilizing a-priori knowledge \cite{relatedwork_eventpred}. The prediction of   events  can be considered as a classification problem in which the probability of a next  event  $a$ given the state $s$ of the process at time $\tau$, $P(a|s(\tau))$, is to be found.

Early predictive models focused on analytical approaches. Le et al. \cite{litrev_le} introduced a hybrid approach consisting of a sequence alignment technique to extract similar patterns and to predict upcoming  events  based on a combination of Markov models. The next  event  of a running process case is therefore determined by the transition probabilities of the Markov models.

Becker et al. \cite{litrev_becker} faced this problem with a similar approach in which historical event data is used to create a Probabilistic Finite Automaton. 
In comparison, Ceci et al. \cite{litrev_ceci} proposed an approach which can handle incomplete traces which is robust to noise and deals with overfitting. This approach leverages sequence mining. Efficient frequent pattern mining is applied to create a tree where prediction models are associated to each node (also called \textit{nested model learning}). These prediction models can be any traditional machine learning algorithms for classification. 

Lakshmanan et al. \cite{litrev_lakshmanan} developed a method which models a process in a probabilistic and instance-specific way. This model can predict next   events  and can be translated into a Markov chain. Their approach has been implemented on a simulated automobile insurance claim process.

Similarly, Unuvar et al. \cite{litrev_unuvar} proposed a method to predict the likelihood of future process tasks by modeling parallel paths which can be either dependent or independent. The authors applied their methodology to a simulated marketing campaign business process model.

More recently, Breuker et al. \cite{litrev_breuker} introduced a predictive model based on the theory of grammatical inference. They have modeled business processes probabilistically with a method called RegPFA which is based on Probabilistic Finite Automaton. Grammatical inference is applied on top of the finite automaton. One of the advantages is that the methodology is based on weaker biases while maintaining comprehensibility. This is important because users without deep technical knowledge can interpret and understand the models. Breuker et al. evaluated their approach against two public available real-world logs demonstrating significant performance improvements. Breuker et al. are able to predict the next event given a running process case with accuracies between $69\%$ and $81\%$ according to their reports.

Most recent research studies have shown the applicability of deep learning to predict process states and events. Evermann et al. \cite{litrev_evermann} have shown in 2017 that recurrent neural networks can be applied to predict next events in processes and improve state-of-the-art prediction accuracies. They create word embeddings from each event instance of the event log to train an LSTM neural network. Therefore, the process is modeled implicitly by the neural network itself. Evermann et al. used the same datasets as Breuker et al. \cite{litrev_breuker} for comparison. 

A comparable approach has been elaborated by Tax et al. \cite{litrev_tax} who predict next events including their timestamps and remaining case times using LSTMs. This approach is similar to the one Evermann et al. demonstrated before. However, according to Khan et al. \cite{litrev_khan}, a major drawback of LSTMs in this context is their limited memory due to the predefined size of the memory state representation which is used to predict next events. They claim that distant event instances in long-running cases vanish over time from the memory state vector. Therefore, Khan et al. \cite{litrev_khan} adapted to overcome the memory limitations of LSTMs by applying memory-augmented neural networks. This technique leverages external memory modules for long-term retention to model complex event processes. The authors demonstrate the applicability and report slight performance improvements compared to Tax et al. \cite{litrev_tax}.

A further approach has been elaborated by Mehdiyev et al. \cite{litrev_mehdiyev}. The authors encode events into n-gram features using a sliding window approach and leverage feature hashing on top. These features, in turn, are used to train a deep learning model consisting of unsupervised stacked autoencoders and supervised fine-tuning. This architecture has shown significant performance improvements across most of the datasets, yet it is more complex compared to the methods described earlier. Mehdiyev et al. can predict the next event given a running process case with accuracies between $66\%$ and $83\%$ according to their reports.

Since deep learning techniques are difficult to interpret, Lee et al. \cite{litrev_lee} developed a method based on matrix factorization and knowledge from business process management to create predictive models which are easier to understand. The authors claim to require fewer parameters than neural networks while maintaining good performance. 

In this work, we have three major contributions.
First, we propose an approach to represent process model states in a continuous rather than a discrete format by enhancing PNs with time decay functions. Second, we show that we can use this approach to incorporate time as a continuous feature to predict the next process  event  since the duration between two events might be correlated with the type of subsequent occurring events in real-world processes. At the same time, we retain a comprehensible process model with its advantages \cite{processmining, interpretability, interpretability2, interpretability3}. With these advancements, we demonstrate that our next  event  prediction algorithm performs significantly better than the previously introduced methods. Third, we contribute a comprehensive evaluation of recent next  event  prediction algorithms across nine benchmark datasets reporting five different classification evaluation metrics.

\section{Preliminaries}\label{sec:prelim}
In this section, we introduce the preliminaries which are required throughout the paper. We introduce event logs, followed by PNs. We provide a general introduction to Process Mining in \ref{sec:process-mining} and introduce a state-of-the-art process discovery algorithm in Section \ref{sec:split-miner}. Finally, we define neural networks in Section \ref{sec:neural-networks}.

\subsection{Event Logs}\label{sec:prelim-el}
The definitions in this subsections are partially  based on the work of van der Aalst et al. \cite{processmining} and Guo et al. \cite{definitions}.

An \textit{event} $a \in \mathcal{A}$ describes an instantaneous change of a process' state. In this definition, $\mathcal{A}$ is the finite set of all possible events. Example events based on the process steps described in Section \ref{sec:introduction} are \textit{start negotiation}, \textit{end fraud assessment}, and \textit{start offer creation}. A specific event $a$ may happen more than once in a given process. An event instance $E$ is a vector with at least two attributes: the name of the associated event $a$ and the corresponding occurrence timestamp. An instance vector may contain further non-mandatory attributes like costs, people, and resources associated to that event occurrence. 
Based on the definition of an event instance, two event instances cannot have the same timestamp, i.e. cannot occur simultaneously. This is because of the continuous nature of time, and the fact that point probabilities in continuous probability distributions are zero.

We define $\mathcal{N}$  as the set of all possible  event instances  and $\mathcal{D}$ as the set of all possible attributes. Then for any event instance $E \in \mathcal{N}$  and any attribute  $d \in \mathcal{D}:\upsilon_d(E)$  is the value of the attribute $d$ for the event instance $E$. If an event instance $E$ does not contain an attribute $d$, then  $\upsilon_d(E) = \varnothing $ (empty set). We denote the attribute \textit{timestamp} by $d_{ts}$. 

A \textit{case} $g$ is a finite and chronological sequence of event instances. 
In literature, the term \textit{trace} is also used to describe a \textit{case}, thus we use both terms synonymously.  We define $\mathcal{G}$ as the finite set of all possible traces and $\gamma(g)$ as a function that returns the number of event instances of a trace $g \in \mathcal{G}$, i.e. the length of $g$. 

An \textit{event log} is a set of traces $\mathcal{L} \subseteq \mathcal{G}$.  Moreover, $\mathcal{L}_{i,j}$ refers to the $j$th event instance in the $i$th trace of an event log $\mathcal{L}$. $|\mathcal{L}|$ denotes the cardinality of $\mathcal{L}$ corresponding to its number of traces. Similarly, $\gamma(\mathcal{L}_{i})$ expresses the number of event instances of the $i$th trace of the event log $\mathcal{L}$. 

\subsection{Petri Net}\label{sec:petri-nets}
A PN is a  mathematical model  that can represent a process. It consists of a set of places; these are graphically represented as circles and transitions represented as rectangles. Transitions correspond to events. Transitions and places are also referred to as \textit{nodes}. Additionally, arcs are used to unidirectionally connect places to transitions and vice versa. A labeled PN is defined as 
\begin{eqnarray}
PN = \langle\mathcal{P}, \mathcal{T}, \mathcal{F}, \mathcal{A}, \pi\rangle
\end{eqnarray}
where $\mathcal{P}$ is the set of places, $\mathcal{T}$ is the set of transitions,  $\mathcal{F} \subseteq (\mathcal{P} \times \mathcal{T}) \cup (\mathcal{T} \times \mathcal{P})$ is the set of directed arcs connecting places and transitions, and $\mathcal{A}$ is the set of events \cite{processmining, definitions, definitions2}. The set $\mathcal{P} \cup \mathcal{T}$ is called the set of nodes. The first node of each pair $(x,y) \in \mathcal{F}$ represents always the source whereas the second node represents always the sink of the directed arc. In other words,  a node $x$ is the input node to another node $y$ iff $(x,y) \in \mathcal{F}$. Similarly,  $x$ is the output node to another node $y$ iff $(y,x) \in \mathcal{F}$. For any $x \in \mathcal{P} \cup \mathcal{T}$, $\bullet x = \{y | (y,x) \in \mathcal{F}\}$ is the set of \textit{input nodes} to $x$ and $x \bullet = \{y | (x,y) \in \mathcal{F}\}$ is the set of \textit{output nodes} of $x$. 
The function $\pi: \mathcal{T} \rightarrow \mathcal{A} \cup \{\perp\}$ maps each transition $t\in\mathcal{T}$ to either a single event of $\mathcal{A}$ or to the non-observable event $\perp$. A labeled PN is defined such that
\begin{eqnarray}
\forall_{a \in \mathcal{A}} \exists ! _{t \in \mathcal{T}} \pi(t) = a.
\end{eqnarray}

Each place can hold a non-negative integer number of tokens. We define $\sigma(p)$ as the number of tokens in a place $p$ where $p \in \mathcal{P}$. 

The state of a PN corresponds to a marking $M \in \mathcal{M}$ where $\mathcal{M}$ is the set of all possible markings. 
We define $M \in Z^{|\mathcal{P}|}$ as a vector of size $|\mathcal{P}|$ where $\mathcal{Z}$ denotes the set of all non-negative integers and $|\mathcal{P}|$ corresponds to the cardinality of $\mathcal{P}$. Each element $M_i = \sigma(p_i), i = 1, ..., |\mathcal{P}|$ where $p_i$ is the $i$th place of $\mathcal{P}$.
The initial state $M^{init}$ is also called \textit{initial marking}, whereas the final state $M^{final}$ is called \textit{final marking} \cite{processmining}.  Usually discovered process models in process mining have a dedicated source and a dedicated sink place that indicate the start and end of the process. All other process nodes are on a path between them. Hence, $M^{init}$ and $M^{final}$ describe the process source and sink states\cite{processmining}.  

Moreover, a transition $t \in \mathcal{T}$ is  mathematically defined as  \textit{enabled}  \cite{processmining} , i.e. can only be fired if
\begin{eqnarray}
\forall_{p \in \bullet t} \sigma(p) \geq 1.
\end{eqnarray}
Hidden transitions, a special type of transition, are associated to the non-observable  event  $\perp$. Such transitions can always fire independent of observed  events  as long as the introduced token requirements at incoming places are met.
When firing a transition $t$, a token is removed from each of the input places $\bullet t$, while a token is added to each of the output places $t \bullet$.

Process models do not always behave as desired. For example, PNs may contain unintended deadlocks or transitions that can never become enabled. Different criteria have been specified under the term \textit{soundness} to prevent process models from such behavior \cite{soundness_aalst2, soundness}. 
It is defined as follows \cite{processmining}. A labeled PN with dedicated source and sink places is considered sound iff: 
\begin{itemize}
  \item for any place $p \in \mathcal{P}$, $p$ cannot hold multiple tokens at the same time,
  \item for any marking $M \in \mathcal{M}$ that indicates a token in the dedicated sink place of the PN, $M = M^{final}$ which implies that there are no remaining tokens in other places than the dedicated sink one when the final marking is reached,
  \item for any marking $M \in \mathcal{M}$, the final marking $M^{final}$ is reachable,
  \item and for any $t \in \mathcal{T}$, a firing sequence of events exists that enables $t$.
\end{itemize}

Furthermore, we define a function $\delta_p(g)$ for all $p \in \mathcal{P}$ measuring the average time between a token leaves a place $p$ until a new token enters $p$ based on an input trace $g$. Finally, $\tau_p$ describes the most recent time that a token entered a place $p$.

\subsection{Process Mining}\label{sec:process-mining}
Process mining defines the discovery, conformance, and enhancement of business processes \cite{processmining, processmining2}. Process discovery is the algorithmic extraction of process models from event logs. One can carry out analysis on obtained models which are usually in the format of PNs, Business Process Modeling Notations (BPMN), Event Driven Process Chains (EPCs), or Casual Nets (CN). In this paper, we will focus on PNs only.

Conformance is defined as the evaluation of the quality of a discovered process model, i.e. if it is a good representation of the process recorded by an event log. It is commonly evaluated based on \textit{fitness} and \textit{precision} among other metrics \cite{processmining}. Therefore, each trace of an event log is replayed by executing the events sequentially on top of the process model. \textit{Fitness} metric functions evaluate the quality of a process model by quantifying deviations between an event log and the replay response of a process model to this event log. A process model should allow replaying the behavior seen in the event log \cite{processmining}. \textit{Precision} metric functions represent the alignment between simulated traces from the obtained process model and true traces from the event log. Ideally, each generated trace by the process model should be realistic, thus being present in the actual event log.

Enhancement considers discovered process models as well as event logs to improve or extend the models. Examples of process enhancement include structural corrections to allow the occurrence of specific behavior or extending a process model with performance data.

\subsection{Split Miner}\label{sec:split-miner}
\textit{Split miner} \cite{splitminer} is a process discovery algorithm  that creates sound labeled PNs with dedicated source and sink places from event logs and that  is characterized by recent significant performance improvements in comparison to existing state-of-the-art methods \cite{pnbenchmark}. It is currently the best algorithm to automatically obtain PN process models from event logs with high fitness and precision. This discovery method has been developed to engage with the tradeoff between fitness, precision, and the complexity of the obtained process model. 

\textit{Split miner} consists of the following five steps \cite{splitminer}. First, it discovers a directly-follows dependency graph and detects short loops. In the second step, the algorithm searches for concurrency and marks the respective elements as such. Afterward, \textit{split miner} applies filtering such that each node is on a path from a single start node to an end node to guarantee \textit{soundness}, the number of edges are minimal to reduce complexity, and that every path from start to end has the highest possible sum of frequencies to maximize fitness. Fourth, the algorithm adds \textit{split} gateways to capture choice and concurrency. As the final step, this discovery method detects \textit{joins}. 

\textit{Split miner} encompasses two hyperparameters: a frequency threshold $\varepsilon$ to control the filtering process and $\eta$ which is a threshold to control parallelism detection. Both hyperparameters are percentiles, i.e. the numerical range is between $0$ and $1$.  Moreover, this algorithm considers only the sequence of events without timestamp or other related information  during process discovery.
The discovery algorithm is publicly available as a Java application \cite{researchcode}.

\subsection{Neural Network}\label{sec:neural-networks}
A neural network is a computing methodology motivated by biological nervous systems. Such networks consist of a set of artificial neurons which receive one or multiple inputs and produce one output. This set is divided into a predefined number of disjoint subsets $n$ where $n \geq 2$. Each subset represents a layer $l_n$ in the form of a matrix containing  outputs of the corresponding neurons. We refer to layer $l_1$ as the input and $l_n$ as the output layer of the neural network. Multiple so-called hidden layers can exist in between. In a fully connected neural network, all neurons of a layer $l_k$ are connected to all neurons of its adjacent layer $l_{k+1}$ for $k \leq n-1$. A very basic neural network can be defined in the following way \cite{neuralnetwork, deeplearningbook}.

\begin{figure*}[ht]
  \begin{center}
    \includegraphics[width=\textwidth]{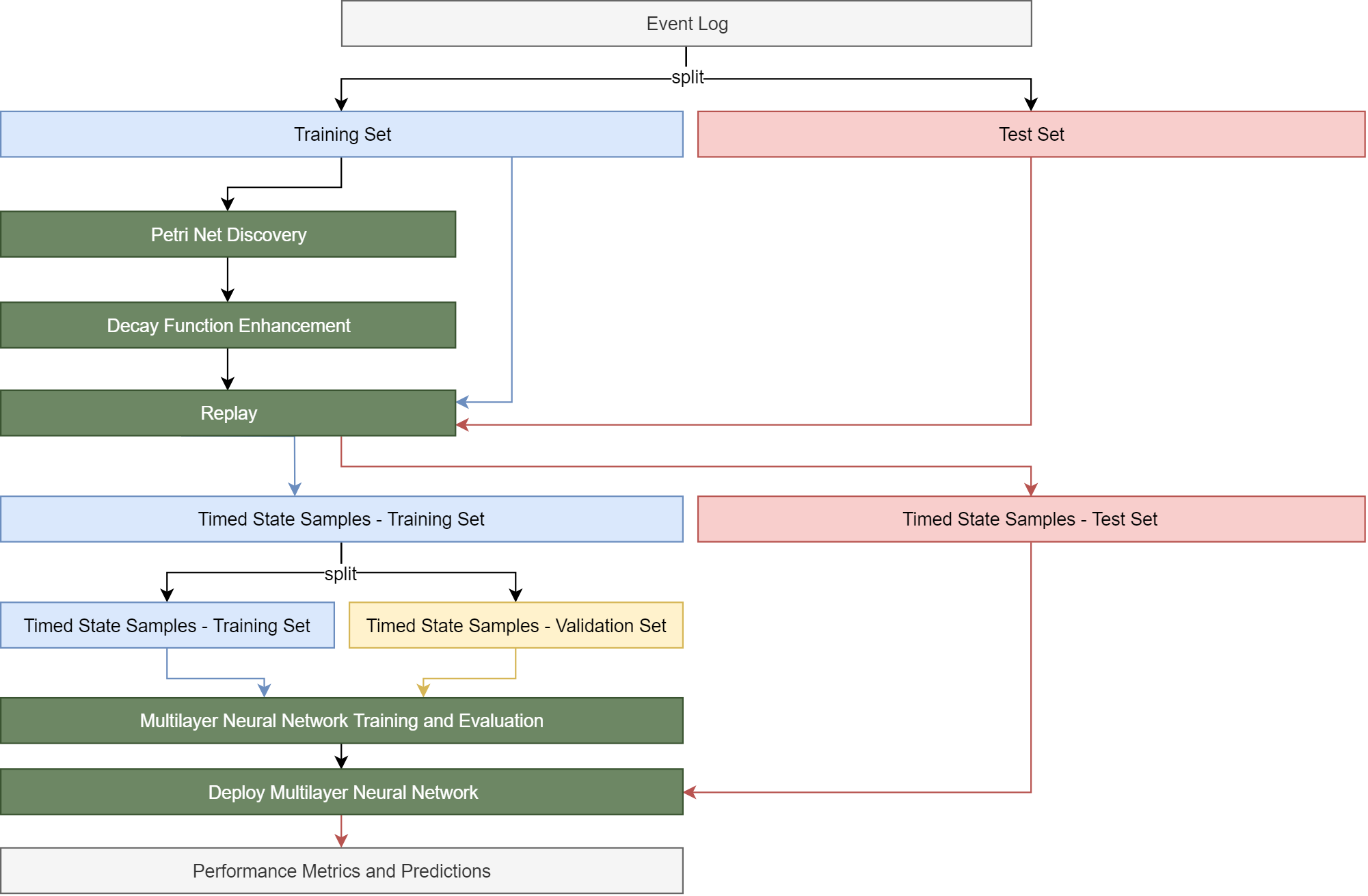}
  \caption{This figure illustrates the flow diagram of the proposed DREAM-NAP approach. It also visualizes the training and testing procedures. The elements of the approach are shown in green, train datasets in blue, test datasets in red, and evaluation datasets in yellow colors. The flows are color-coded correspondingly.}
  \label{fig:algo_flow}
  \end{center}
\end{figure*}
A neuron $j$ which belongs to layer $l_{k}$ calculates its output based on the weighted outputs of each predecessor neuron of layer $l_{k-1}$. Each direct connection between two neurons $i$ and $j$ is associated with a weight $w_{i,j}$. 
Each neuron $j$ comprises a differentiable activation function $\rho_j$ which is used to calculate the output of a neuron. Thus, the output of a neuron $j$ belonging to $l_{k}$ based on its predecessor layer $l_{k-1}$ can be calculated as 
\begin{eqnarray}
\theta_j(l_{k-1}) = \rho_j(\phi_j(l_{k-1})).
\end{eqnarray}
It follows that
\begin{eqnarray}
\phi_j(l_{k-1}) = \sum_{i}  \theta_{i}(l_{k-2}) * w_{i,j}+w_{0,j}
\end{eqnarray}
where $w_{0,j}$ is a bias term. Such a neural network is commonly modeled as an optimization problem where a cost function $\xi$ is to be defined as a function of the difference between neural network outputs and true values and to be minimized by adapting the weights $w$ of the neural network. This is called a supervised learning problem \cite{deeplearningbook}. 

\section{Approach}\label{sec:approach}
The DREAM-NAP approach  is a supervised learning algorithm to predict the next event given a partial trace of an event log. The method  consists of three steps. First, we discover a PN model from an event log and associate each place of the PN with a decay function. Then, we replay the event log used for discovery and extract feature arrays incorporating decay function response values, token movement counts, and utilized resources. Finally, we train a neural network to predict the  next event  based on these feature arrays. A flow diagram of the training and testing procedure of our approach is visualized in Figure \ref{fig:algo_flow}. In this section, we introduce each component in detail.

The source code of the proposed approach is available in our GitHub repository \footnote{ https://github.com/ProminentLab/DREAM-NAP}.

\subsection{Decay Function Enhancement}\label{sec:enhancement}
To discover a PN, the corresponding event log has to consist of at least one non-empty trace. 
We draw on an existing PN discovery algorithm called \textit{split miner} which has been introduced in Section \ref{sec:split-miner}.

Decay functions are used to model data values that decrease over time. Such functions are commonly applied to population trend modeling, financial domains, and physical systems. The basic form of a decay function is
\begin{eqnarray}\label{eq:decayfunction}
f(\tau) = \beta - \alpha*\tau
\end{eqnarray}
where $\tau$ is time, $\alpha$ is the rate of decay, and $\beta$ is a constant corresponding to the initial value. The decay function $f(\tau)$ can be easily modified to model more complex behavior such as exponential or squared declines. However, the linear decay function presented in Equation \ref{eq:decayfunction} is the simplest option.

We leverage the properties of these functions to expand discovered PN models by decaying the activation of places over time. A place activation is triggered through token arrivals during an event log replay, i.e. when the marking of a PN changes. In this way, we add a continuous time-based dimension to the discrete state representation of PNs. This approach overcomes limitations of the state-of-the-art next  event  prediction methods. In particular, such a decay mechanism leads to the incorporation of time as a continuous variable and can be used to detect and model event instance interarrival times through continuous state representations. At the same time, we preserve all advantages of an interpretable process model \cite{processmining, interpretability, interpretability2, interpretability3} in the form of PNs. Subsequently, we introduce the detailed process of decay function enhancement. 

We associate each place of the PN generated by \textit{split miner} on an event log $\mathcal{L}$ with a linear decay function $f_p(\tau)$. We denote the time difference between the current time, $\tau$, and the most recent time a token has entered place $p$, $\tau_p$, by $\Delta_p = \tau - \tau_p$. 
\begin{eqnarray}\label{eq-decay}
\begin{aligned}
f_p(\tau) = &
\begin{cases}
\beta - \alpha * (\tau - \tau_p) &\text{if } \tau - \tau_p < \beta / \alpha,\\
0 &\text{if otherwise}.
\end{cases}\\
= &
\begin{cases}
\beta - \alpha * \Delta_p~~~~~~~~~ &\text{if } \Delta_p < \beta / \alpha,~~~~~\\
0 &\text{if otherwise}.
\end{cases}
\end{aligned}
\end{eqnarray}

We initially set $\Delta_p = \infty$  for all $p \in \mathcal{P}$ such that $f_p(\tau) = 0$. In this way, we reset all decay functions of the PN. A decay function $f_p(\tau)$ will \textit{activate} as soon as a token enters a corresponding place $p$. The value of this function declines over time $\tau$ and \textit{reactivates} with a response value $\beta$ immediately when a token enters this place. 

During replay, each event instance of an event log corresponds to a transition which fires immediately when a respective  event  is observed and token requirements are met. Instead of focusing on the fired transitions itself, we can also unambiguously identify the sequence of fired transitions by observing the movement of tokens between places. By enhancing each place with a decay function described in Equation \ref{eq-decay}, we assign a level of importance to recent token movements compared to past ones. This mechanism scales event time information into a range from $0$ to $\beta$ without discretization and loss of generality. 

We control the level of importance using the two decay function parameters $\beta$ and $\alpha$. Ideally, $\alpha$ should be set such that the slope of $f_p(\tau)$ covers the whole range from $\beta$ to $0$ based on the \textit{reactivation} durations of a place $p$. In other words, the slope should not be too steep such that $f_p(\tau) = 0$ for a small $\Delta_p$, nor too flat such that $f_p(\tau) \approx \beta$ for a large $\Delta_p$. This cannot be achieved using a single $\alpha$ value for all decay functions of the PN when applying this mechanism to real-world processes with varying durations of reactivation. For this reason, we estimate an individual decay rate $\alpha_p$ for each place $p \in \mathcal{P}$. We define the set of all decay rates as $\mathcal{R}$ where the cardinality of $\mathcal{R}$, $|\mathcal{R}|$, equals to $|\mathcal{P}|$.

We estimate $\alpha_p$ by utilizing the event log $\mathcal{L}$ and the respective PN discovered by \textit{split miner} on $\mathcal{L}$. Each trace $g \in \mathcal{L}$ consists of a finite number of event instances. We refer to the $j$th event instance  of the $i$th trace of an event log $\mathcal{L}$ by $\mathcal{L}_{i,j}$, as mentioned in Section \ref{sec:prelim-el}. The maximum trace duration observed in $\mathcal{L}$ is denoted by $\Delta_{max}(\mathcal{L})$ and is defined with the following equation.

\begin{eqnarray}
\begin{split}
\Delta_{max}(\mathcal{L}) = max\Big(\forall_{1 \leq i\leq |\mathcal{L}|}\big(\upsilon_{d_{ts}}(\mathcal{L}_{i,\gamma(\mathcal{L}_{i})}) - \\ \upsilon_{d_{ts}}(\mathcal{L}_{i,1})\big)\Big)
\end{split}
\end{eqnarray}

For the estimation of $\alpha_p$, it is inevitable to know if a value for $\Delta_p$ exists, i.e. if a place $p$ gets activated only once or if reactivations occur. Therefore, we define a function $\nu_p(g)$ which returns the number of tokens that enter a place $p$ when replaying a trace $g$. We estimate  $\alpha_p$ for two different cases based on the outcome of the following condition.
\begin{eqnarray}\label{eq:deltaLexists}
max\big(\forall_{g \in \mathcal{L}} \nu_p(g)\big) \leq 1
\end{eqnarray}

If Condition \ref{eq:deltaLexists} holds, $\alpha_p$ will be set to a value such that the response of $f_p(\tau)$ will never equal to $0$ before the last event  instance  of a corresponding trace $g \in \mathcal{L}$ occurred. By doing so, we guarantee to carry information on the occurrence of a specific event in the response of the decay function until the end of a trace. Equation \ref{eq:alpapsingle} defines $\alpha_p$ mathematically for this case.
\begin{eqnarray}\label{eq:alpapsingle}
\alpha_p(\mathcal{L}) = \frac{\beta}{\Delta_{max}(\mathcal{L})}
\end{eqnarray}

If Condition \ref{eq:deltaLexists} does not hold, we consider the average reactivation duration of a place $p$ based on all traces of the respective event log. With this information, we set the decay rate to a value such that $f_p(\tau)$ provides a level of recent token movement importance for the average duration between reactivations. Consequently, the slope will neither be too steep nor too flat.
Mathematically, we can estimate $\alpha_p$ by
\begin{eqnarray}
\alpha_p(\mathcal{L}) = \frac{\beta}{mean\big(\forall_{g \in \mathcal{L}} \delta_p(g)\big)}
\end{eqnarray}
where $mean(\cdot)$ is the arithmetic mean function. 

\subsection{Event Log Replay}\label{sec:EventLogReplay}
After estimating all $\alpha_p$ of $\mathcal{R}$ for each place $p$ in $\mathcal{P}$, we can use the corresponding decay functions, $f_p(\tau)$, to obtain a decay function response for all $p$ at a specific time $\tau$. We write $F(\tau)$ as the vector of decay function response values. Each element of this vector corresponds to the response value of one specific place in the PN, i.e. the $i$th element in $F(\tau)$ corresponds to the response of the decay function at time $\tau$ associated to the $i$th place in $\mathcal{P}$. 

Since $F(\tau)$ constitutes only the most recent activation of $\mathcal{P}$, we introduce a counting vector $C(\tau)$ of size $|\mathcal{P}|$ elements where the $i$th element corresponds to the $i$th place in $\mathcal{P}$. We initialize the counting vector at time $0$, $C(\tau = 0)$ by setting each element to $0$. When a token enters a specific place $p$ at time $\tau$, the corresponding counter element will be incremented by $1$ such that $C(\tau)$ reflects the number of tokens which have entered each place from time $0$ to $\tau$. 

Similarly, we introduce a counting vector $R(\tau)$ which counts the occurrence of each unique  non-mandatory event instance attribute value  from time $0$ to $\tau$ when replaying an event log. Continuous attribute values require discretization in advance. 

We replay the event log $\mathcal{L}$ on the PN which has been enhanced using decay functions. $F(\tau)$, $C(\tau)$, $R(\tau)$, and the PN marking $M$ at time $\tau$, $M(\tau)$, will be reset before a trace $g \in \mathcal{L}$ will be replayed. We then obtain vectors and PN states at each time $\tau$ corresponding to the timestamp values of the replayed  event instances  in $\mathcal{L}$. 
 We concatenate the vectors of decay function values, token and resource counts, and the PN marking at time $\tau$ to obtain a single vector that can be used to train a neural network. 
This concatenation of $F(\tau)$, $C(\tau)$, $R(\tau)$, and $M(\tau)$ is called a \textit{timed state sample} $S(\tau)$, 
\begin{eqnarray}
S(\tau) = F(\tau) \oplus C(\tau) \oplus M(\tau) \oplus R(\tau) 
\end{eqnarray}
where $\oplus$ represents a vector concatenation. Therefore, a \textit{timed state sample} $S(\tau)$ describes a PN process state in a timed manner through decay function enhancement. It contains information about time-based token movements, i.e. when a token has entered a place the last time relative to the current time, token counts per place (loop information), and the current PN state using the marking. Optionally, if event instances of the event log contain non-mandatory attributes, the \textit{timed state sample} also contains such information. 

After replaying the event log $\mathcal{L}$, we obtain a set of  \textit{timed state samples}, $\mathcal{S}$, such that Condition \ref{eq:timedpnstatesamples1} holds. 
\begin{eqnarray}\label{eq:timedpnstatesamples1}
\forall_{1 \leq i\leq |\mathcal{L}|} \forall_{1 \leq j\leq \gamma(\mathcal{L}_i)} : S\big(\upsilon_{d_{ts}}(\mathcal{L}_{i,j})\big) \in \mathcal{S}
\end{eqnarray}

\subsection{Deep Learning}\label{sec:approach-dl}
We use the set of  \textit{timed state samples}, $\mathcal{S}$, to predict the next  event.
For each $S_i \in \mathcal{S}$ where $1 \leq i \leq |S|$, we predict the next  event $a$ of the upcoming instance  $E_{i+1}$ given that the  \textit{timed state sample}  $S_i$ does not contain the final marking $M^{final}$. This is a supervised classification problem as the event log $\mathcal{L}$ and the set of  events  $\mathcal{A}$ are known.
An event log $\mathcal{L}$ usually consists of thousands of event instances  across multiple traces. Hence, a deep neural network is a suitable method to conquer this problem due to a large amount of available data.

We propose two fully connected neural network architectures. One which ignores attribute value count vectors $R(\tau)$ in $S_i$, and another one which considers each $S_i$ as is. 
With \textit{DREAM-NAP}, we refer to the first neural network architecture, whereas \textit{DREAM-NAPr} refers to the second one considering event attributes. 
The details of the architecture for \textit{DREAM-NAP} are illustrated in Table \ref{table:nn_fc_arch} whereas the details of \textit{DREAM-NAPr} are illustrated in Table \ref{table:nn_fcr_arch}. Both architectures have been developed in Python using Keras \cite{keras} with a Tensorflow backend \cite{tensorflow}.

\begin{table}
\resizebox{\textwidth}{!}{%
    \centering
    \begin{tabular}{|c|c|}
    \hline
    \textbf{Parameter}     & \textbf{Value} \\ \hline
    \# layers             &    5    \\ \hline
    \# neurons per layer           &  $[$input, input*1.2, input*0.6, input*0.3, output$]$ \\ \hline
    \# dropout layer  &  4    \\ \hline
    dropout rate &    0.2               \\ \hline
    \# batch normalization layers    &   0  \\ \hline
    activation functions             &   $[$relu, relu, relu, relu, softmax$]$    \\ \hline
    loss            & categorical crossentropy  \\ \hline
    optimizer     &    adam                \\ \hline
    \end{tabular}
}
\caption{Deep learning architecture of the \textit{DREAM-NAP} model}
\label{table:nn_fc_arch}
\end{table}

The \textit{DREAM-NAP} neural network consists of five layers. The first layer has the same size as the vector length of $3 * |\mathcal{P}|$ and correspondingly called \textit{input}. The second layer has $1.2$ times, the third $0.6$ times, and the fourth $0.3$ times the size of the \textit{input} layer. Each of these layers use Rectified Linear Unit (ReLU) activation functions, which have proven major performance advantages over sigmoid and hyperbolic tangent ones  in deep learning architectures \cite{relu}. Since the proposed architecture is shallow and traditional neural network activation functions can perform well on such architectures \cite{sigmoid1, sigmoid2, sigmoid3, sigmoid4}, we examined the impact of ReLU and sigmoid activation functions on the predictive accuracy. In all our experimental cases, ReLU-based architectures performed with a higher or equal accuracy score compared to sigmoidal ones. Hence, we propose a ReLU-based \textit{DREAM-NAP} architecture.  The final layer is the output layer with a size equal to $|\mathcal{A}|$. 

The output layer utilizes a softmax activation function since we are interested in the probability of a specific $a \in \mathcal{A}$. We use dropout \cite{dropout} for regularization applied between each hidden layer as well as between the fourth and the output layer. We decide on the Adam optimizer \cite{adam} to train the neural network. Batch normalization \cite{batchnorm} layers are not used in this architecture since no further regularization is required. Moreover, batch normalization did not improve the results of the \textit{DREAM-NAP} architecture, as we will demonstrate in Section \ref{sec:evaluation}.

\begin{table}
\resizebox{\textwidth}{!}{
    \centering
    \begin{tabular}{|c|c|}
    \hline
    \textbf{Parameter}     & \textbf{Value} \\ \hline
    \# layers             &    6    \\ \hline
    \# neurons per layer           &  ~~~~~~~~~~~$[$input, 300, 200, 100, 50, output$]$~~~~~~~~~~~~~~\\ \hline
    \# dropout layer  &  5    \\ \hline
    dropout rate &    0.5               \\ \hline
    \# batch normalization layers    &    5 \\ \hline
    activation functions             &   \makecell{$[$relu, relu, relu, relu, relu, softmax$]$\\ 
    $[$sigm, sigm, sigm, sigm, sigm, softmax$]$
    }\\ \hline
    loss            & categorical crossentropy  \\ \hline
    optimizer     &    adam                \\ \hline
    \end{tabular}
}
\caption{Deep learning architecture of the \textit{DREAM-NAPr} model with an activation function hyperparameter}
\label{table:nn_fcr_arch}
\end{table}

The \textit{DREAM-NAPr} architecture is similar to the \textit{DREAM-NAP} one. 
However, in this architecture, we use fixed layer sizes that are a result of a comprehensive grid search over the number of layers and number of neurons per layer. For the architecture search, we considered three, four, and five layers with each either 50, 100, 150, 200, 300, 400, or 500 neurons per layer. The results of this search have shown that the \textit{DREAM-NAPr} model with the architecture described in Table \ref{table:nn_fcr_arch} performs best across all benchmark training datasets. We specifically propose a fixed number of neurons per layer since a dynamic assignment based on the size of $R(\tau)$ could easily result in an unreasonably large number of neurons.

Since this architecture is most likely confronted with a higher probability of overfitting due to the number of event instance  attribute values, we increase the dropout rate and consider batch normalization layers.
Furthermore, the type of neural network activation function is a hyperparameter of this model. We examined the impact of ReLU over sigmoid activation functions on this architecture with inconclusive results. In half of our experimental cases, \textit{DREAM-NAPr} architectures with sigmoid functions performed better whereas in the other half ReLU lead to higher accuracy scores. As a consequence, the choice of activation function type is application-specific and therefore not suggested to be fixed.

\section{Evaluation}\label{sec:evaluation}
In this section, we evaluate our proposed approach using the \textit{DREAM-NAP} and \textit{DREAM-NAPr} models introduced in Section \ref{sec:approach-dl} on nine popular benchmark datasets and compare to the most recent peer-reviewed methods in the literature. We contrast our method specifically to the algorithms of Tax et al. \cite{litrev_tax}, Evermann et al. \cite{litrev_evermann}, Breuker et al. \cite{litrev_breuker}, and Lee et al \cite{litrev_lee}. The source codes of these methods are publicly available. Hence, we evaluate and perform all experiments on the same dataset splits.
Unfortunately, a fair comparison to the method of Mehdiyev et al. \cite{litrev_mehdiyev} is not possible since the authors of that paper did not disclose the corresponding source code and deep learning parameter sets that were required to reproduce the results. 
We, therefore, exclude this method from our statistical comparison.

We first provide an overview of the datasets, followed by the introduction of metrics we will use.  We then report and comment on the conformance of the discovered PN process models of all datasets. Afterward, we describe the preprocessing steps of the  timed state samples  before feeding them to our proposed deep learning architectures. 
Finally, we evaluate the prediction performance of the neural network models. 

We perform the discovery of PNs using \textit{split miner} and the transformation of event logs to  timed state samples  using our \textit{DREAM} approach on a computer running Windows 10 with an Intel i7-6700 CPU and 16GB RAM. This task took between 30 minutes and 4 hours depending on the size of the dataset. The training of the \textit{DREAM-NAP} and \textit{DREAM-NAPr} neural networks were performed on Tesla K80 and NVidia GeForce RTX 2080 Ti GPUs and took between 15 minutes and 2 hours per dataset.

\subsection{Datasets}
Our evaluation is based on three real-life benchmark datasets, specifically the Helpdesk \cite{data_helpdesk}, the Business Process Intelligence Challenge 2012 (BPIC12) \cite{data_bpic12}, and the Business Process Intelligence Challenge 2013 (BPIC13) \cite{data_bpic13} dataset.

The Helpdesk dataset comprises events from a ticketing management process of an Italian software company. Each event instance contains the mandatory event type  and associated timestamp. No further attributes are used.

The BPIC12 dataset originates from a Dutch financial institute and represents the process of a loan application. It can be split into three subprocesses related to the \textit{work}, the \textit{application} itself, and the \textit{offer}. All event instances contain the required attributes as well as further non-mandatory  resource information. Moreover, each event instance  describes a \textit{lifecycle} status which is either \textit{complete}, \textit{scheduled} or \textit{start}. Finally, the event instances of this event log  carry information about the requested loan amount. We split the dataset into multiple subprocesses to be able to compare our results to the results of existing methods. We consider the complete event log without any filtering, denoted by \textit{BPIC12 - all}. \textit{BPIC12 - all complete} considers only event instances  of \textit{lifecycle} value \textit{complete}. Similarly, we filter the original event log by \textit{work} related  events  only and consider all events, code-named as \textit{BPIC12 - work all}, and events with \textit{lifecycle} attribute value \textit{complete} as \textit{BPIC12 - work complete}.
Additionally, we consider the subprocesses of \textit{offer}s and \textit{application}s separately as \textit{BPIC12 - O} and \textit{BPIC12 - A}  to perform our evaluation on the same datasets as the state-of-the-art methods \cite{litrev_evermann, litrev_breuker, litrev_lee, litrev_mehdiyev}. These subprocess event logs consist of event instances  with \textit{complete} lifecycle values only.

The third log originates from Volvo IT and describe events from an incident and problem management system. Each event instance contains the required attributes and non-mandatory information about the lifecycle, group,  responsible employee, resource country, organization country, involved organizations, impact, and the product. Events that are associated with a problem rather than an incident contain a further attribute which describes the role of the affected organization. We split this dataset into two separate event logs handling incidents and problems independently. We call these two event logs \textit{BPIC13 - Incidents} and \textit{BPIC13 - Problems}.

An overview of all datasets  and their  number of  event instances, events, traces, and resource attributes is given in Table \ref{table:dataset_overview}.

\begin{table}[ht]
\resizebox{\textwidth}{!}{%
    \centering
    \begin{tabular}{|c|c|c|c|c|}
    \hline
    \textbf{Dataset}       & \textbf{\# event instances} & \textbf{\# events} & \textbf{\# traces} & \textbf{\# resources} \\ \hline
    Helpdesk               & 13,710             & 9   & 3,804              & 0                     \\ \hline
    BPIC12 - all           & 262,200            & 24  & 13,087             & 2                     \\ \hline
    BPIC12 - all complete  & 164,506            & 23  & 130,897            & 2                     \\ \hline
    BPIC12 - work complete & 72,413             & 6   & 9,658              & 2                     \\ \hline
    BPIC12 - work all      & 170,107            & 7   & 9,658              & 2                     \\ \hline
    BPIC12 - O             & 31,244             & 7   & 5,015              & 2                     \\ \hline
    BPIC12 - A             & 60,849             & 10  & 13,087             & 2                     \\ \hline
    BPIC13 - Incidents     & 65,533             & 4   & 7,554              & 7                     \\ \hline
    BPIC13 - Problems      & 8,599              & 4    & 1,758              & 8                     \\ \hline
    \end{tabular}
    \caption{Number of event instances,  events, traces, and resources for each of the evaluated datasets.}
    \label{table:dataset_overview}
}
\end{table}

\subsection{Metrics}
We utilize 10-fold cross-validation to perform our evaluation. Therefore, we consider $90\%$ of the actual traces for training and $10\%$ for testing. The training set is used to discover multiple process models, as we describe in Section \ref{sec:pndiscovery}.
We measure the quality of the obtained PNs using a basic conformance checking function called \textit{token-based replay fitness} (Equation \ref{eq:fitness}), adopted from \cite{processmining}. The function calculates the fitness by replaying each trace of an event log based on the number of missing, consumed, remaining, and produced tokens. The higher its score, the higher the alignment between the event log and the process model. Such fitness functions are common process mining metrics to evaluate the quality of process models \cite{rozinat, rozinat2, fitnesseval, fitnesseval2}, as described in Section \ref{sec:process-mining}.
\begin{eqnarray}\label{eq:fitness}
fitness = \frac{(1-\frac{missing}{consumed})}{2} + \frac{(1 - \frac{remaining}{produced})}{2}
\end{eqnarray}
We select the best fitting process model after process discovery based on the fitness function of Equation \ref{eq:fitness}. This model is then used for decay function enhancement and estimation of the corresponding decay function parameters, as described in Section \ref{sec:enhancement}. 

Afterward, the enhanced PN model is used to replay the training as well as the test set to obtain  timed state samples. Moreover, we split the training set after replaying into a $90\%$ training and $10\%$ holdout evaluation set. We finally obtain three disjoint datasets for training, validation, and testing a deep learning model. We train deep learning models on the training set only and select the best one based on the validation set. The best predictive model is chosen at the lowest validation loss which is an effective and widely used approach to train neural networks called early stopping \cite{earlystopping1, earlystopping2, earlystopping3}. An overview of this procedure is visualized in Figure \ref{fig:algo_flow}.

We evaluate the predictive performance of our approach based on averaged accuracy, precision, and recall, as well as F-score and the \textit{area under the curve} (AUC) of the receiver operating characteristic. All of the metrics are used to compare against the earlier-introduced next event prediction techniques. The subsequent definitions of metrics are based on \cite{litrev_mehdiyev, measures, auc}.

\textit{Accuracy} is defined as 
\begin{eqnarray}
\frac{1}{|\mathcal{S}|}\sum_{i=1}^{|\mathcal{A}|} n_i * \frac{tp_i + tn_i}{tp_i + tn_i + fp_i + fn_i}
\end{eqnarray}
where $|\mathcal{S}|$ is the total number of  timed state samples  and $n_i$ the number of  timed state samples  with a next  event  equal to the $i$th  event  in $\mathcal{A}$. Moreover, $tp$, $tn$, $fp$, and $fn$ represent \textit{true positive}, \textit{true negative}, \textit{false positive}, and \textit{false negative} respectively. 

\textit{Precision} is defined as 
\begin{eqnarray}
\frac{1}{|\mathcal{S}|}\sum_{i=1}^{|\mathcal{A}|} n_i * \frac{tp_i}{tp_i + fp_i}.
\end{eqnarray}

\textit{Recall} is defined as
\begin{eqnarray}
\frac{1}{|\mathcal{S}|}\sum_{i=1}^{|\mathcal{A}|} n_i * \frac{tp_i}{tp_i + fn_i}.
\end{eqnarray}

In addition, we report the \textit{F-score} for each dataset. This measure is the harmonic mean of precision and recall and provides information on how precise and robust an algorithm is. \textit{F-Score} is defined as 
\begin{eqnarray}
    \frac{1}{|\mathcal{S}|}\sum_{i=1}^{|\mathcal{A}|} n_i * \frac{precision * recall}{precision + recall}.
\end{eqnarray}

\begin{figure*}[ht]
  \begin{center}
    \includegraphics[width=\textwidth]{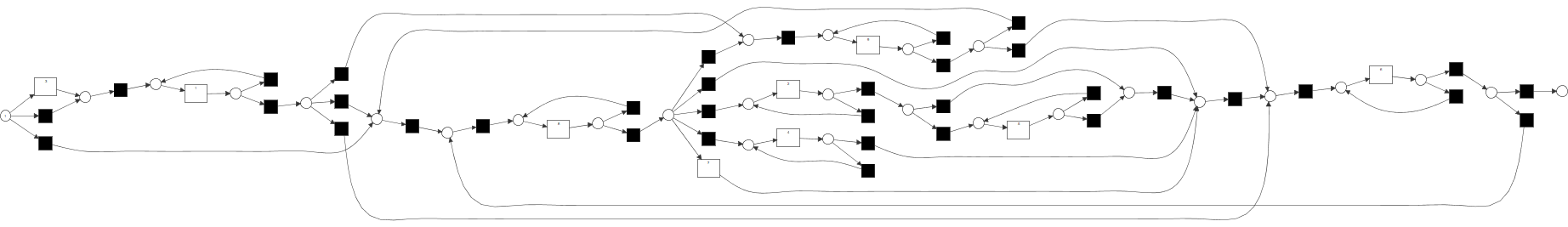}
  \caption{This figure shows the interpretable PN obtained from the first training set of the Helpdesk dataset.}
  \label{fig:helpdesk_pn}
  \end{center}
\end{figure*}

Finally, we report the AUC of the receiver operating characteristic. It is a common classification analysis to determine which model predicts classes best. The closer an AUC value is to $1$, the better the model is. \textit{Multiclass AUC} is defined as 
\begin{eqnarray}
\frac{1}{|\mathcal{S}|}\sum_{i=1}^{|\mathcal{A}|} n_i *  \int_{0}^{1} tpr_i d(fpr_i)
\end{eqnarray}
where $tpr_i$ and $fpr_i$ is the true positive and false positive rate for the $i$th  event.

In a final step, we compare the overall performance of our approach with the ones of the state-of-the-art algorithms using a rank test. Also, we perform a sign test to determine statistical significant improvements. This test method is a variation of a binomial test and considers the number of times an algorithm performed best \cite{stat_comp}.

\subsection{Petri Net Discovery}\label{sec:pndiscovery}
We initially utilize \textit{split miner} to discover multiple PN process models for all benchmark event logs. We perform hyperparameter optimization to obtain the best combination of  $\varepsilon$ and $\eta$ for each of the $10$-fold cross-validation training sets of each dataset. $\varepsilon$ and $\eta$ are initially set to $0.0$ and are increment in $0.1$ steps. A PN is discovered for each of the $100$ hyperparameter combinations. Based on Equation \ref{eq:fitness}, for each fold, we select the process model with the highest fitness score for decay function enhancement. The closer the value is to $1$, the better the PN represents the logical behavior of an underlying process. This logic does not have to be learned by a neural network.

Table \ref{table:pn_perf} illustrates the averaged fitness scores over the best process models for all 10 folds per each benchmark training dataset. It can be seen that \textit{split miner} can detect PNs with fitness values above $85\%$. The models obtained on \textit{BPIC12 - work all}, \textit{BPIC12 - A}, and \textit{BPIC13 - Incidents} even reach fitness scores above $95\%$. This shows that process discovery techniques can unveil and model basic logical behavior from event logs which we leverage in our enhancement approach. However, none of the process model evaluations result in a perfect fitness score of $1$. This is because \textit{split miner} filters infrequent behavior, i.e. discards information, which does not seem to correspond to the main process behavior.

We visualize the best obtained PN from the first fold of the Helpdesk training set in Figure \ref{fig:helpdesk_pn}. The white rectangles represent the $9$  events recorded in the event log, whereas black rectangles correspond to hidden transitions, i.e. transitions which are  mapped to the non-observable event $\perp$. The basic behavior of the underlying process can be observed, interpreted, and analyzed in different contexts. This underscores one of the advantages of process models.
\begin{table}[h]
\resizebox{\textwidth}{!}{%
    \centering
    \begin{tabular}{|c|c|}
    \hline
    \textbf{~~~~~~~~~~~~~~~~~Dataset~~~~~~~~~~~~~~~~~}     & \textbf{~~~~~~~~~Averaged Fitness~~~~~~~~~} \\ \hline
    Helpdesk             &   0.928     \\ \hline
    BPIC12 - all             &    0.919    \\ \hline
    BPIC12 - all complete             &       0.88 \\ \hline
    BPIC12 - work complete             &  0.892      \\ \hline
    BPIC12 - work all             &  0.961      \\ \hline
    BPIC12 - O             &   0.856     \\ \hline
    BPIC12 - A             &   0.951     \\ \hline
    BPIC13 - Incidents             &  0.955      \\ \hline
    BPIC13 - Problems             &  0.925      \\ \hline                            
    \end{tabular}
}
\caption{This table shows the averaged cross validated fitness scores of the PN models obtained from each dataset. It can be seen that \textit{split miner} is able to capture basic logical process behavior for all benchmark event logs.}
\label{table:pn_perf}
\end{table}

\subsection{Deep Learning Preprocessing}
After selecting the best process model for each fold of all benchmark datasets, we enhance these PNs using decay functions and replay the training and testing traces to create  timed state samples, as described in Section \ref{sec:approach} and as visualized in Figure \ref{fig:algo_flow}. However, several preprocessing steps are necessary before feeding the  timed state samples  to the proposed \textit{DREAM-NAP} and \textit{DREAM-NAPr} deep learning architectures.

All datasets originating from \textit{BPIC12} contain one continuous and one categorical resource attribute. We discretize the continuous attribute by quantizing its values using disjoint intervals of size $20$.

For the event logs originating from the \textit{BPIC13} dataset, we consider only the categorical attributes of \textit{resource country}, \textit{organization country}, \textit{involved organization}, \textit{impact}, and, if applicable, \textit{role of the affected organization}. The number of unique values for the excluded resources are too large, hence these resources do not contribute beneficial and generalizable information.

We normalize each component of the  timed state samples  of the training and validation set separately, i.e. $F(\tau)$, $C(\tau)$, $M(\tau)$ and $R(\tau)$, to zero mean and unit variance. The mean and standard deviation of each vector before normalization is used to normalize the test set.

\subsection{Results}\label{sec:results}
\begin{table*}[!ht]
\resizebox{\textwidth}{!}{%
\centering
\begin{tabular}{|c|c|c|c|c|c|c|c|}
\hline
\textbf{Dataset} & \textbf{Approach} & \textbf{Model} & \textbf{Accuracy} & \textbf{Precision} & \textbf{Recall} & \textbf{F-Score}  & \textbf{AUC}\\ \hline
                  & \cellcolor[HTML]{C0C0C0}DREAM & \cellcolor[HTML]{C0C0C0}DREAM-NAP   & \cellcolor[HTML]{C0C0C0}\textbf{0.829} & \cellcolor[HTML]{C0C0C0} 0.768 & \cellcolor[HTML]{C0C0C0}\textbf{0.829} & \cellcolor[HTML]{C0C0C0}\textbf{0.795} & \cellcolor[HTML]{C0C0C0}\textbf{0.876} \\ \cline{2-8} 
                & Tax et al. \cite{litrev_tax}                               &                              & 0.814                                 & \textbf{0.799}                                      & 0.814                                      & 0.792                                      & 0.872                                      \\ \cline{2-2} \cline{4-8} 
                & Evermann et al. \cite{litrev_evermann}                              &                              & 0.599                                  & 0.625                                      & 0.599                                      & 0.550                                      & 0.716                                      \\ \cline{2-2} \cline{4-8} 
                & Breuker et al. \cite{litrev_breuker}                              &                              & 0.801                                  & 0.759                                      & 0.801                                      & 0.777                                      & 0.861                                      \\ \cline{2-2} \cline{4-8} 
 \multirow{-5}{*}{Helpdesk $^*$}                & Lee et al. \cite{litrev_lee}                              &                              & 0.801                                  & 0.772                                      & 0.801                                      & 0.782                                      & 0.861                                      \\ \cline{2-2} \cline{4-8} 
\hline                                         
& \cellcolor[HTML]{C0C0C0}DREAM & \cellcolor[HTML]{C0C0C0}DREAM-NAP   & \cellcolor[HTML]{C0C0C0}0.847          & \cellcolor[HTML]{C0C0C0}0.838          & \cellcolor[HTML]{C0C0C0}0.847          & \cellcolor[HTML]{C0C0C0}0.820          & \cellcolor[HTML]{C0C0C0}0.910          \\ \cline{2-8} 
                                         & \cellcolor[HTML]{C0C0C0}DREAM & \cellcolor[HTML]{C0C0C0}DREAM-NAPr & \cellcolor[HTML]{C0C0C0}\textbf{0.896} & \cellcolor[HTML]{C0C0C0}\textbf{0.895} & \cellcolor[HTML]{C0C0C0}\textbf{0.896} & \cellcolor[HTML]{C0C0C0}\textbf{0.888} & \cellcolor[HTML]{C0C0C0}\textbf{0.942} \\ \cline{2-8} 
                & Tax et al. \cite{litrev_tax}                               &                              & 0.850                                 & 0.848                                      & 0.850                                      & 0.834                                      & 0.917                                      \\ \cline{2-2} \cline{4-8} 
                & Evermann et al. \cite{litrev_evermann}                              &                              & 0.789                                  & 0.801                                      & 0.789                                      & 0.774                                      & 0.885                                      \\ \cline{2-2} \cline{4-8} 
                & Breuker et al. \cite{litrev_breuker}                              &                              & 0.770                                  & 0.698                                      & 0.770                                      &0.722                                      & 0.871                                      \\ \cline{2-2} \cline{4-8} 
\multirow{-5}{*}{BPIC12 - all}                 & Lee et al. \cite{litrev_lee}                              &                              & 0.695                                  & 0.730                                      & 0.695                                      & 0.694                                      & 0.835                                      \\ \cline{2-2} \cline{4-8} 
\hline
                                         & \cellcolor[HTML]{C0C0C0}DREAM & \cellcolor[HTML]{C0C0C0}DREAM-NAP   & \cellcolor[HTML]{C0C0C0}0.789          & \cellcolor[HTML]{C0C0C0}0.778          & \cellcolor[HTML]{C0C0C0}0.789          & \cellcolor[HTML]{C0C0C0}0.746          & \cellcolor[HTML]{C0C0C0}0.884          \\ \cline{2-8} 
                                         & \cellcolor[HTML]{C0C0C0}DREAM & \cellcolor[HTML]{C0C0C0}DREAM-NAPr & \cellcolor[HTML]{C0C0C0}\textbf{0.863} & \cellcolor[HTML]{C0C0C0}\textbf{0.871} & \cellcolor[HTML]{C0C0C0}\textbf{0.863} & \cellcolor[HTML]{C0C0C0}\textbf{0.856} & \cellcolor[HTML]{C0C0C0}\textbf{0.926} \\ \cline{2-8} 
                & Tax et al. \cite{litrev_tax}                               &                              & 0.802                                 & 0.794                                      & 0.802                                      & 0.767                                      & 0.892                                      \\ \cline{2-2} \cline{4-8} 
                & Evermann et al. \cite{litrev_evermann}                              &                              & 0.684                                  & 0.693                                      & 0.684                                      & 0.645                                      & 0.829                                      \\ \cline{2-2} \cline{4-8} 
                & Breuker et al. \cite{litrev_breuker}                              &                              & 0.721                                  & 0.639                                      & 0.721                                      & 0.654                                      & 0.847                                      \\ \cline{2-2} \cline{4-8} 
 \multirow{-6}{*}{BPIC12 - all complete}               & Lee et al. \cite{litrev_lee}                              &                              & 0.681                                  & 0.722                                      & 0.681                                      & 0.678                                      & 0.830                                      \\ \cline{2-2} \cline{4-8} 
\hline
                                         & \cellcolor[HTML]{C0C0C0}DREAM & \cellcolor[HTML]{C0C0C0}DREAM-NAP   & \cellcolor[HTML]{C0C0C0}\textbf{0.747}          & \cellcolor[HTML]{C0C0C0}\textbf{0.748 }         & \cellcolor[HTML]{C0C0C0}\textbf{0.747}          & \cellcolor[HTML]{C0C0C0}\textbf{0.720}          & \cellcolor[HTML]{C0C0C0}\textbf{0.827 }         \\ \cline{2-8} 
                                         & \cellcolor[HTML]{C0C0C0}DREAM & \cellcolor[HTML]{C0C0C0}DREAM-NAPr & \cellcolor[HTML]{C0C0C0}\textbf{0.823}          & \cellcolor[HTML]{C0C0C0}\textbf{0.823} & \cellcolor[HTML]{C0C0C0}\textbf{0.823}          & \cellcolor[HTML]{C0C0C0}\textbf{0.818}          & \cellcolor[HTML]{C0C0C0}\textbf{0.872  }        \\ \cline{2-8} 
                & Tax et al. \cite{litrev_tax}                               &                              & 0.698                                 & 0.732                                      & 0.698                                      & 0.688                                      & 0.804                                      \\ \cline{2-2} \cline{4-8} 
                & Evermann et al. \cite{litrev_evermann}                              &                              & 0.709                                  & 0.723                                      & 0.709                                      & 0.693                                      & 0.805                                      \\ \cline{2-2} \cline{4-8} 
 \multirow{-5}{*}{BPIC12 - work complete $^{**}$}               & Lee et al. \cite{litrev_lee}                              &                              & 0.547                                  & 0.526                                      & 0.547                                      & 0.524                                      & 0.689                                      \\ \cline{2-2} \cline{4-8} 
\hline
                                         & \cellcolor[HTML]{C0C0C0}DREAM & \cellcolor[HTML]{C0C0C0}DREAM-NAP   & \cellcolor[HTML]{C0C0C0}\textbf{0.878 }         & \cellcolor[HTML]{C0C0C0}\textbf{0.882 }         & \cellcolor[HTML]{C0C0C0}\textbf{0.878}          & \cellcolor[HTML]{C0C0C0}\textbf{0.877}          & \cellcolor[HTML]{C0C0C0}\textbf{0.919 }         \\ \cline{2-8} 
                                         & \cellcolor[HTML]{C0C0C0}DREAM & \cellcolor[HTML]{C0C0C0}DREAM-NAPr & \cellcolor[HTML]{C0C0C0}\textbf{0.902} & \cellcolor[HTML]{C0C0C0}\textbf{0.903} & \cellcolor[HTML]{C0C0C0}\textbf{0.902} & \cellcolor[HTML]{C0C0C0}\textbf{0.900} & \cellcolor[HTML]{C0C0C0}\textbf{0.934} \\ \cline{2-8} 
                & Tax et al. \cite{litrev_tax}                               &                              & 0.855                                 & 0.875                                      & 0.855                                      & 0.861                                      & 0.909                                      \\ \cline{2-2} \cline{4-8} 
                & Evermann et al. \cite{litrev_evermann}                              &                              & 0.855                                 & 0.876                                     & 0.855                                      & 0.861                                      & 0.908                                      \\ \cline{2-2} \cline{4-8} 
                & Breuker et al. \cite{litrev_breuker}                              &                              & 0.850                                  & 0.856                                      & 0.850                                      & 0.848                                      & 0.901                                      \\ \cline{2-2} \cline{4-8} 
 \multirow{-6}{*}{BPIC12 - work all}               & Lee et al. \cite{litrev_lee}                              &                              & 0.784                                  & 0.784                                      & 0.784                                      & 0.778                                      & 0.862                                      \\ \cline{2-2} \cline{4-8} 
\hline
                                         & \cellcolor[HTML]{C0C0C0}DREAM & \cellcolor[HTML]{C0C0C0}DREAM-NAP   & \cellcolor[HTML]{C0C0C0}\textbf{0.852}          & \cellcolor[HTML]{C0C0C0}0.888          & \cellcolor[HTML]{C0C0C0}\textbf{0.852}          & \cellcolor[HTML]{C0C0C0}\textbf{0.834 }         & \cellcolor[HTML]{C0C0C0}\textbf{0.918}          \\ \cline{2-8} 
                                         & \cellcolor[HTML]{C0C0C0}DREAM & \cellcolor[HTML]{C0C0C0}DREAM-NAPr & \cellcolor[HTML]{C0C0C0}\textbf{0.935} & \cellcolor[HTML]{C0C0C0}\textbf{0.932} & \cellcolor[HTML]{C0C0C0}\textbf{0.935} & \cellcolor[HTML]{C0C0C0}\textbf{0.927} & \cellcolor[HTML]{C0C0C0}\textbf{0.963} \\ \cline{2-8} 
                & Tax et al. \cite{litrev_tax}                               &                              & 0.837                                 & 0.876                                      & 0.837                                      & 0.824                                      & 0.910                                      \\ \cline{2-2} \cline{4-8} 
                & Evermann et al. \cite{litrev_evermann}                              &                              & 0.807                                  & 0.876                                      & 0.807                                      & 0.797                                      & 0.893                                      \\ \cline{2-2} \cline{4-8} 
                & Breuker et al. \cite{litrev_breuker}                              &                              & 0.826                                  & 0.894                                      & 0.826                                      & 0.820                                      & 0.905                                      \\ \cline{2-2} \cline{4-8} 
  \multirow{-6}{*}{BPIC12 - O}              & Lee et al. \cite{litrev_lee}                              &                              & 0.822                                  & 0.826                                      & 0.822                                      & 0.812                                      & 0.900                                      \\ \cline{2-2} \cline{4-8} 
\hline
                                         & \cellcolor[HTML]{C0C0C0}DREAM & \cellcolor[HTML]{C0C0C0}DREAM-NAP   & \cellcolor[HTML]{C0C0C0}\textbf{0.805}          & \cellcolor[HTML]{C0C0C0}\textbf{0.748}          & \cellcolor[HTML]{C0C0C0}\textbf{0.805}          & \cellcolor[HTML]{C0C0C0}\textbf{0.761}          & \cellcolor[HTML]{C0C0C0}\textbf{0.893}          \\ \cline{2-8} 
                                         & \cellcolor[HTML]{C0C0C0}DREAM & \cellcolor[HTML]{C0C0C0}DREAM-NAPr & \cellcolor[HTML]{C0C0C0}\textbf{0.951} & \cellcolor[HTML]{C0C0C0}\textbf{0.958} & \cellcolor[HTML]{C0C0C0}\textbf{0.951} & \cellcolor[HTML]{C0C0C0}\textbf{0.950} & \cellcolor[HTML]{C0C0C0}\textbf{0.974} \\ \cline{2-8} 
                & Tax et al. \cite{litrev_tax}                               &                              & 0.791                                 & 0.739                                      & 0.791                                      & 0.746                                      & 0.885                                      \\ \cline{2-2} \cline{4-8} 
                & Evermann et al. \cite{litrev_evermann}                              &                              & 0.601                                  & 0.640                                      &  0.601                                     & 0.530                                      & 0.765                                      \\ \cline{2-2} \cline{4-8} 
                & Breuker et al. \cite{litrev_breuker}                              &                              & 0.800                                  & 0.745                                      & 0.800                                      & 0.755                                      & 0.890                                      \\ \cline{2-2} \cline{4-8} 
\multirow{-6}{*}{BPIC12 - A}                             & Lee et al. \cite{litrev_lee}                              &                              & 0.805                                  & 0.748                                      & 0.805                                      & 0.761                                      & 0.893                                      \\ \cline{2-2} \cline{4-8} 
\hline
                                         & \cellcolor[HTML]{C0C0C0}DREAM & \cellcolor[HTML]{C0C0C0}DREAM-NAP   & \cellcolor[HTML]{C0C0C0}\textbf{0.703}          & \cellcolor[HTML]{C0C0C0}0.689         & \cellcolor[HTML]{C0C0C0}\textbf{0.703}          & \cellcolor[HTML]{C0C0C0}0.657          & \cellcolor[HTML]{C0C0C0}0.685          \\ \cline{2-8} 
                                         & \cellcolor[HTML]{C0C0C0}DREAM & \cellcolor[HTML]{C0C0C0}DREAM-NAPr & \cellcolor[HTML]{C0C0C0}\textbf{0.882} & \cellcolor[HTML]{C0C0C0}\textbf{0.879} & \cellcolor[HTML]{C0C0C0}\textbf{0.882} & \cellcolor[HTML]{C0C0C0}\textbf{0.879} & \cellcolor[HTML]{C0C0C0}\textbf{0.880} \\ \cline{2-8} 
                & Tax et al. \cite{litrev_tax}                               &                              & 0.697                                 & 0.708                                      & 0.697                                      & 0.667                                      & 0.699                                      \\ \cline{2-2} \cline{4-8} 
                & Evermann et al. \cite{litrev_evermann}                              &                              & 0.637                                  & 0.642                                      & 0.637                                      & 0.619                                      & 0.667                                      \\ \cline{2-2} \cline{4-8} 
                & Breuker et al. \cite{litrev_breuker}                              &                              & 0.697                                  & 0.709                                      & 0.697                                      & 0.661                                      & 0.686                                      \\ \cline{2-2} \cline{4-8} 
\multirow{-6}{*}{BPIC13 - Incidents}                 & Lee et al. \cite{litrev_lee}                              &                              & 0.637                                  & 0.655                                      & 0.637                                      & 0.618                                      & 0.714                                      \\ \cline{2-2} \cline{4-8} 
\hline
                                         & \cellcolor[HTML]{C0C0C0}DREAM & \cellcolor[HTML]{C0C0C0}DREAM-NAP   & \cellcolor[HTML]{C0C0C0}\textbf{0.675}          & \cellcolor[HTML]{C0C0C0}0.604          & \cellcolor[HTML]{C0C0C0}\textbf{0.675}          & \cellcolor[HTML]{C0C0C0}\textbf{0.593}          & \cellcolor[HTML]{C0C0C0}0.558          \\ \cline{2-8} 
                                         & \cellcolor[HTML]{C0C0C0}DREAM & \cellcolor[HTML]{C0C0C0}DREAM-NAPr & \cellcolor[HTML]{C0C0C0}\textbf{0.791} & \cellcolor[HTML]{C0C0C0}\textbf{0.777} & \cellcolor[HTML]{C0C0C0}\textbf{0.791} & \cellcolor[HTML]{C0C0C0}\textbf{0.779} & \cellcolor[HTML]{C0C0C0}\textbf{0.779} \\ \cline{2-8} 
                & Tax et al. \cite{litrev_tax}                               &                              & 0.630                                 & 0.608                                      & 0.630                                      & 0.584                                      & 0.591                                      \\ \cline{2-2} \cline{4-8} 
                & Evermann et al. \cite{litrev_evermann}                              &                              & 0.624                                  & 0.654                                      & 0.624                                      & 0.565                                      & 0.568                                      \\ \cline{2-2} \cline{4-8} 
 \multirow{-5}{*}{BPIC13 - Problems $^{**}$}               & Lee et al. \cite{litrev_lee}                              &                              & 0.597                                  & 0.593                                      & 0.597                                      & 0.570                                      & 0.627                                      \\ \cline{2-2} \cline{4-8} 
\hline
\end{tabular}%
}
\caption{This table illustrates the results obtained by the proposed approach and contrasts them to existing state-of-the-art methods. Bold values designate that the proposed model outperforms state-of-the-art results. $^*$ denotes datasets that do not contain resources, therefore \textit{DREAM-NAPr} is not applicable. $^{**}$ denotes that the source code of Breuker et al. \cite{litrev_breuker} was not able to produce results on this dataset.}
\label{table:results}
\end{table*}

\begin{figure*}[ht]
  \begin{center}
    \includegraphics[width=\textwidth]{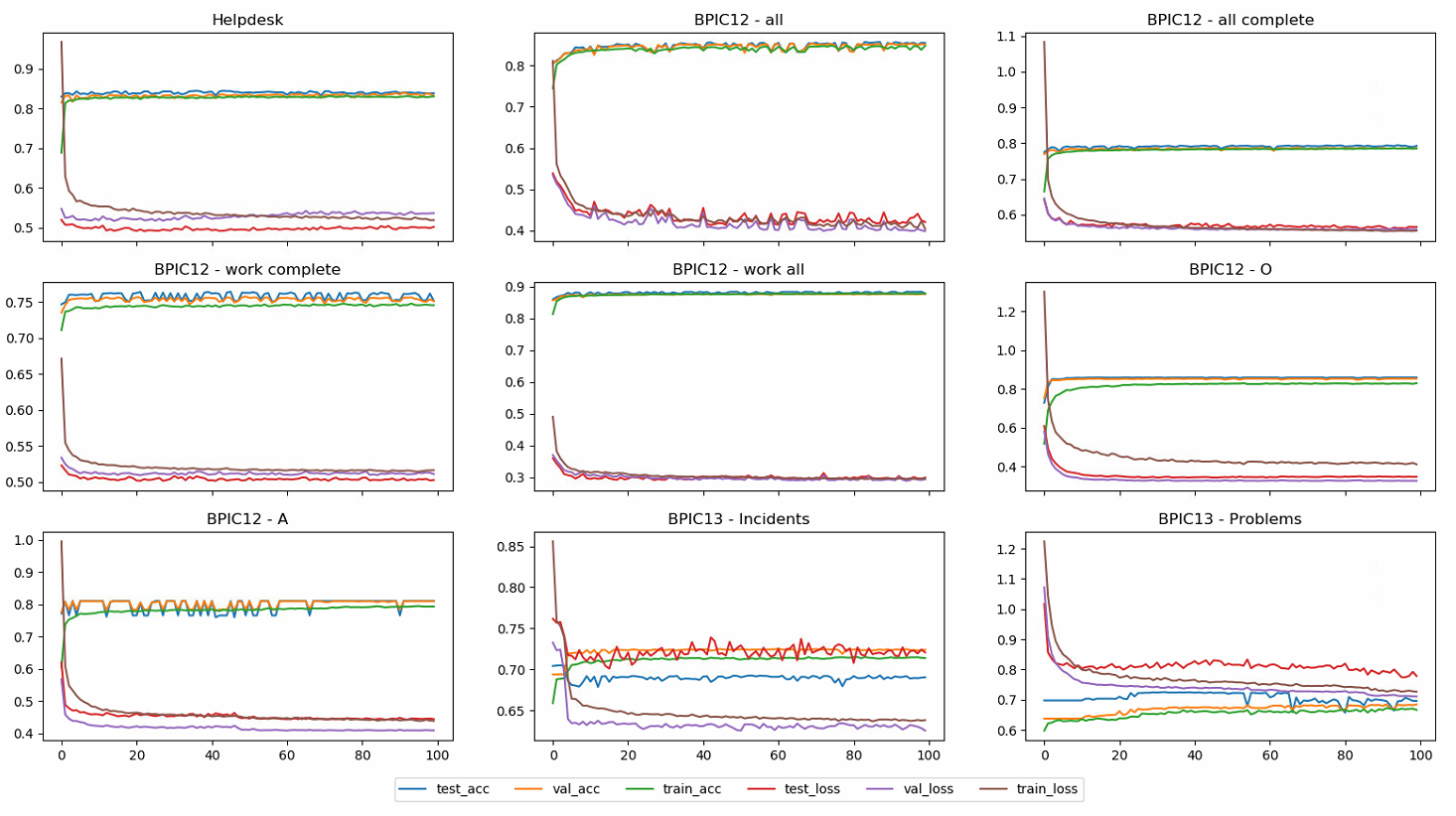}
  \caption{This figure shows the training, test, and validation accuracy and loss (y-axis) over 100 training epochs (x-axis) for each dataset without considering  non-mandatory event attributes. Each plot shows the first cross-validation run representative for all ten runs.}
  \label{fig:eval_fc_plots}
  \end{center}
\end{figure*}
\begin{figure*}[ht]
  \begin{center}
    \includegraphics[width=\textwidth]{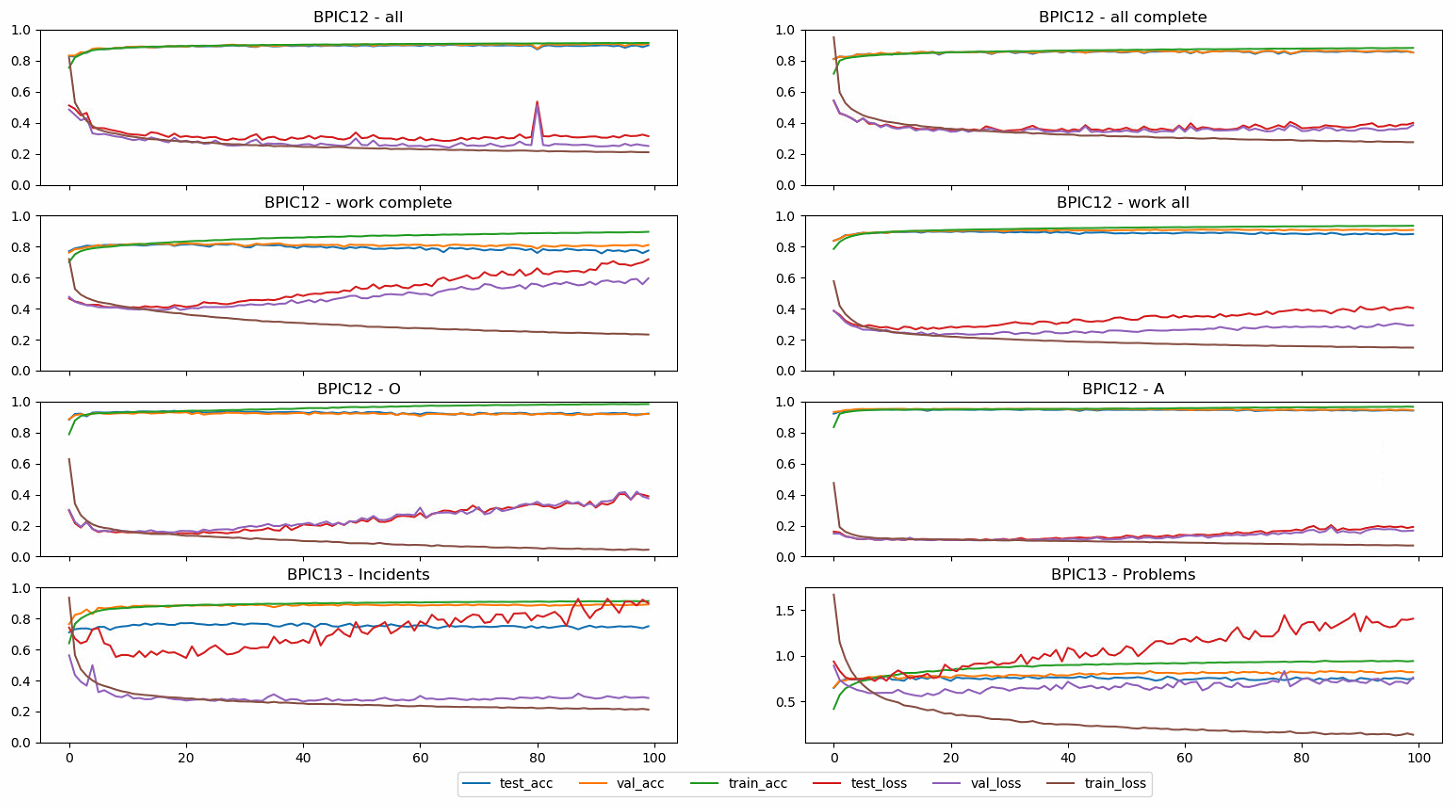}
  \caption{This figure shows the training, test, and validation accuracy and loss (y-axis) over 100 training epochs (x-axis) for each dataset  with utilized non-mandatory event attributes. \textit{BPIC12 - work complete}, \textit{BPIC12 - O}, \textit{BPIC13 - Incidents} and \textit{BPIC13 - Problems} start overfitting comparatively early. Each plot shows the first cross-validation run representative for all ten runs.}
  \label{fig:eval_fcr_plots}
  \end{center}
\end{figure*}
We train the \textit{DREAM-NAP} and \textit{DREAM-NAPr} models on the preprocessed training timed state samples  of each benchmark dataset  using early stopping, but continue training for in total $100$ epochs for visualization purposes. The training batch size of \textit{DREAM-NAP} is set to $64$ whereas the \textit{DREAM-NAPr} batch size is set to $100$ to accelerate the training process. We train the methods of Evermann et al. \cite{litrev_evermann}, Tax et al. \cite{litrev_tax}, Breuker et al. \cite{litrev_breuker}, and Lee et al. \cite{litrev_lee} using the reported sets of hyperparameters. However, the methods of Evermann et al., Tax et al., and Lee et al. are not designed to predict the first event of a sequence. According to the corresponding authors, the inputs of the methods must be padded with null/zero  events  such that the first predictable  event  equals the actual first  event  of the trace. In this way, all methods can be applied and evaluated on the same datasets. The detailed results are listed in Table \ref{table:results}.

\textit{DREAM-NAP} outperforms seven out of nine benchmark datasets in terms of accuracy and recall, six out of nine in terms of F-score, and five out of nine in terms of AUC. Especially on the \textit{BPIC12 - work complete} dataset, we demonstrate that the decay mechanism in combination with token movement counts is extremely beneficial to predict the next  event. We outperform the current state-of-the-art by $3.8\%$ in accuracy and recall as well as with a $2.5\%$ higher precision.  This leads to an F-Score value of $72\%$. The state-of-the-art method with the closest F-Score value is Evermann et al. \cite{litrev_evermann} with $69.3\%$. Lee et al. \cite{litrev_lee} perform the worst on this dataset with an F-Score of $52.4\%$. This underscores that our approach performs significantly better on this dataset than existing state-of-the-art methods. 
\textit{DREAM-NAP} also surpasses the existing methods on \textit{Helpdesk}, \textit{BPIC12 - work all}, \textit{BPIC12 - O}, \textit{BPIC13 - Incidents}, and \textit{BPIC12 - Problems} in terms of accuracy and recall. However, the improvement is less significant.  On the \textit{Helpdesk} dataset, the F-Score of \textit{DREAM-NAP} is only $0.3\%$ greater than the state-of-the-art obtained by Tax et al. \cite{litrev_tax}. Furthermore, the method of Evermann et al. \cite{litrev_evermann} results in the lowest observed F-Score value with $55\%$ on this dataset, though performing very well on \textit{BPIC12 - work complete}. This indicates that not all state-of-the-art methods perform consistently well across all datasets. 
For the \textit{BPIC12 - A} dataset, we obtain  slightly better scores compared  to the state-of-the-art across all observed metrics. Our proposed \textit{DREAM-NAP} model does not outperform the existing methods when considering all process  events  of the BPIC12 dataset, i.e. on \textit{BPIC12 - all complete} and \textit{BPIC12 - all}, though performing with scores close to the state-of-the-art. We show that our approach achieves a comparatively stable performance across the evaluated benchmark datasets, whereas other state-of-the-art methods perform more unstable, such as the earlier mentioned method of Evermann et al. \cite{litrev_evermann}. In terms of precision, one observes that \textit{DREAM-NAP} only outperforms the state-of-the-art in one-third of the datasets. It often falls to the LSTM-based approaches of Tax et al. \cite{litrev_tax} and Evermann et al. \cite{litrev_evermann}. Future research is suggested to increase the precision of the \textit{DREAM-NAP} model  to reduce the false positive rate. In five out of nine cases, our approach outperforms the state-of-the-art in terms of F-score and AUC. For all \textit{BPIC12} datasets, the obtained F-scores values are in the range of $72\%$ to $87\%$ describing satisfactory performance, but leaving room for improvement. Especially when analyzing the \textit{BPIC13} datasets, one observes F-score values of $65.7\%$ and $59.3\%$ which are comparatively good, but which also leave room for improvements. Accordingly, the AUC values for these datasets leave opportunities for further enhancements, too. 

Ultimately, \textit{DREAM-NAP} scores  consistently  average to high ranks without considering resource information. This underscores that PNs extended with decay functions and token movement counters carry important information to predict the next  event  in running process cases. However, we also see that further research should be conducted to improve the quality of our predictions, especially in terms of precision.

The \textit{DREAM-NAPr} architecture outperforms the state-of-the-art in terms of accuracy, precision, recall, F-score, and AUC on all eight out of eight datasets containing event resource information. Similar to the \textit{DREAM-NAP} model, the slightest improvements are observed on the \textit{BPIC12} datasets that consider all  types of events. In these two cases, we outperform the state-of-the-art by $4.6\%$ and $6.3\%$ in accuracy and recall. At the same time, we improve the precision on these datasets resulting in higher and more desirable F-score values. The AUC scores of $94.2\%$ and $92.6\%$ indicate strong and worthwhile classification results.
The results of the \textit{BPIC12} subprocesses show accuracy, precision, and recall uptakes between $3.8\%$ and $21\%$. Especially the results on \textit{BPIC12 - work complete} and \textit{BPIC12 - A} show that the incorporation of  event  resource information can dramatically increase the predictive performance. As a result, the F-score and AUC values for these datasets are much higher than the ones of the state-of-the-art  indicating consistent results with desired well-balanced false positive and false negative rates. 
In the same way, the results on the \textit{BPIC13} datasets show significant improvements between $12.3\%$ and $18.5\%$ in terms of accuracy, precision, and recall. A large amount of available resource information contains critical information to predict the next event. Although we are improving the overall performance, predicting the next  event  for \textit{BPIC13 - Problems} remains difficult. None of our reported metric scores are greater than $80\%$ leaving space for further enhancement.

Overall, it can be seen that the predictive performance of the proposed approaches is significantly larger compared to the existing methods. Moreover, our models perform with well balanced scores across all benchmark datasets resulting in comparatively better F-score and AUC values. Solely \textit{DREAM-NAP} on \textit{Helpdesk}, \textit{BPIC12 - A}, and \textit{BPIC13 - Problems} has a $6.1\%$, $5.7\%$, and $7.1\%$ lower precision compared to its accuracy and recall.

Figure \ref{fig:eval_fc_plots} shows the training, evaluation, and validation accuracy and loss over $100$ epochs of the \textit{DREAM-NAP} architecture for each dataset. It can be seen that none of the models tend to overfit. This confirms that batch normalization layers are not required for this neural network architecture. All models demonstrate a smooth learning curve and converge after a few training epochs.

Figure \ref{fig:eval_fcr_plots} visualizes the same metrics scores over training epochs for the \textit{DREAM-NAPr} models. In comparison to the previous figure, all datasets tend to overfit early. Especially on \textit{BPIC12 - work complete}, our architecture overfits and demonstrates the importance of early stopping.  It can be noted that the models which overfit the earliest and strongest, are the models that do not improve much compared to the state-of-the-art. Specifically, \textit{DREAM-NAPr} on \textit{BPIC12 - work complete} shows strong overfitting. At the same time, this model is more than $17\%$ below a perfect accuracy. Similarly, overfitting can be observed on \textit{BPIC13 - Incidents} and \textit{BPIC13 - Problems}; two further models that result in low outperforming accuracies in comparison to all benchmarks.

\begin{figure}[h!]
  \begin{center}
    \includegraphics[width=\textwidth]{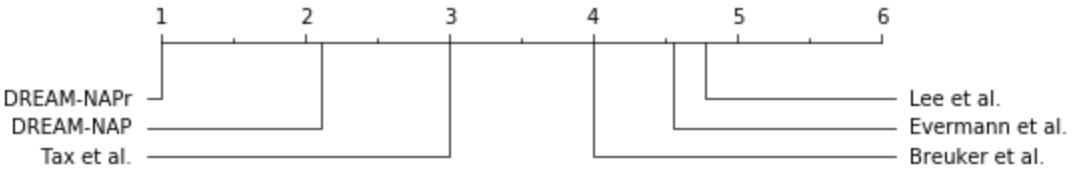}
  \caption{Arithmetic means of ranks of the state-of-the-art and proposed approaches.}
  \label{fig:crit_diff}
  \end{center}
\end{figure}

The diagram shown in Figure \ref{fig:crit_diff} indicates the superiority of our proposed architectures in terms of accuracy. \textit{DREAM-NAP} scores an average arithmetic rank of $2.1$ whereas \textit{DREAM-NAPr} scores on average first. In comparison, the method proposed by Tax et al. \cite{litrev_tax}, which performs with competitive scores across all metrics and datasets, and which beats the \textit{DREAM-NAP} model in accuracy in two of the datasets, scores an average rank of $3$. 

We can further statistically test whether the improvements in accuracy of our proposed approach are significant by comparing our architectures against the best state-of-the-art algorithm on each dataset. We are using a sign test due to the small number of available samples, i.e. $9$ for \textit{DREAM-NAP} and $8$ for \textit{DREAM-NAPr}. We set the level of significance to $\alpha = 0.05$ and adjust it using the Dunn-Sidak correction \cite{sidak} to control the type I error. Therefore, the level of significance for \textit{DREAM-NAP} is
\begin{eqnarray}
\alpha_{nap} = 1 - (1 - 0.05)^{1/9} = 0.0057
\end{eqnarray}
 and for \textit{DREAM-NAPr} is 
 \begin{eqnarray}
\alpha_{napr} = 1 - (1 - 0.05)^{1/8} = 0.0064.
\end{eqnarray}
The sign test for \textit{DREAM-NAP} results in a $p$-value of $0.0898$ whereas it results in a value smaller than $0.00001$ for \textit{DREAM-NAPr}. We can see that $0.00001 \leq 0.0064$, thus \textit{DREAM-NAPr} shows significant improvements in accuracy over the state-of-the-art. This  further  underscores the superiority of our proposed method.

The results show that the \textit{DREAM-NAP} model performs with consistently high metric scores across a diverse set of event logs without considering  event resource information. The performance is further improved when incorporating available non-mandatory attributes  using \textit{DREAM-NAPr}. Hence, we can deduce that the proposed \textit{DREAM} approach adds significant value to the deep learning predictor. 
Overall, we demonstrate statistical superiority over the state-of-the-art methods, therefore presenting major improvements.

\section{Conclusion}\label{sec:conclusion}
In this paper, we introduced a novel approach to predict next events in running process cases called \textit{DREAM-NAP}. Specifically, we extended the places of PN process models with decay functions to obtain  timed state samples  when replaying an event log. These timed samples are used to train a deep neural network which accurately predicts the next event in a running process case. Our results surpass many state-of-the-art techniques. We obtain cross-validated accuracies above $90\%$ and show robust, precise performances across a diverse set of real-world event logs. This underscores the feasibility and usefulness of our proposed approach.

We have shown that decay functions are a suitable tool to express a traditionally discrete PN state as a continuous representation during process runtime. In this way, we can incorporate timing information of processes directly into the process model. This is important for predictive tasks such as predicting the next event  since the duration between two event instances might be correlated with a subsequent occurring  event.

While most recent techniques model processes implicitly, our approach is based on explicit process models. Therefore, our method is easier to interpret than algorithms which are based  exclusively  on deep learning. While decision making of neural networks is naturally hard to understand and explain \cite{conclusion_xai, conclusion_xai2, conclusion_xai3}, we are retaining an interpretable process model  in combination with a simple deep learning architecture. Therefore, organizations will still be able to debug their processes using graphical representations of PNs while taking advantage of the predictive capabilities. A sensitivity analysis can be performed to interpret the decision making of the neural network which performs on top of the decay function extended PN process model.

This paper introduced a promising novel approach with many potential real-world applications. High quality next event predictions are beneficial for the efficient control of real-time processes. Predicting future events helps organizations to improve scheduling, model flexible demand, and reduce system waste which are high impact problems to tackle issues like climate change \cite{climatechange}. Other applications can be found in atypical process mining disciplines such as in black box controller logic estimation \cite{theis2019process} that has the potential to release huge amounts of engineers from heavy-duty of repeated controller design work. The proposed approach can further be applied in the novel process mining discipline of Human-Computer Interaction \cite{Theis:2019:BPN:3340630.3331155}. Predicting upcoming user interactions with a given computer system might unveil error-prone or inefficient interfaces and can be used to create accessible and interactive devices to overcome e.g. user impairments \cite{Wobbrock:2019:SAM:3319499.3330292}. Finally, DREAM-NAP might have potential applications in Healthcare to understand patient's medical records with the ultimate goal to optimize treatments and diagnoses. 

Further research can be conducted in the following three directions. 
First, the predictive quality might be able to be improved by incorporating quality performance measures of process discovery algorithms apart from fitness scores to  investigate the impact of the process model quality on the proposed approach. Additionally, repair methods might be beneficial to increase the process quality and may have a positive impact on the predictive performance \cite{repair1, repair2, repair3}.  The outcome of such studies might further increase the metric scores on the next event prediction that we have reported. Second, we have applied the simplest kind of decay function. A comprehensive study of different decay function types might improve the predictive performance of our approach. Moreover, we proposed two deep learning architectures that have shown satisfying results on the evaluated benchmark datasets. Further optimized architectures might exist that increase the quality of predicting next  events  and that might overcome the low precision scores reported in Section \ref{sec:results}. Finally, the presented approach has been applied to next event prediction only, but might apply to  further  predictive process management tasks such as remaining case time prediction, next event timestamp prediction, or anomalous process state predictions. 
\newpage

\section*{Appendix: Notations}
\begin{description}[style =standard, labelindent=0em , labelwidth=2.5cm, labelsep*=1em, leftmargin =!]
\item [$\varnothing$] empty set 
\item [$\perp$ ] \textit{non-observable event}
\item [$a$] \textit{ event }
\item [$\mathcal{A}$] finite set of all \textit{ events }
\item [$\alpha$] decay rate
\item [$\alpha_p$] decay rate for a specific place $p$
\item [$\beta$] constant parameter of a decay function 
\item [$C(\tau)$] token counting vector from time $0$ to $\tau$, each element represents the number of tokens which entered a specific place
\item [$d$] attribute
\item [$d_{ts}$] timestamp attribute
\item [$\mathcal{D}$] finite set of all possible attributes
\item [$\delta_p(g)$] function of average time between a token is consumed in place $p$ until a new token is produced in $p$ based on an input trace $g$
\item [$\Delta_p$] time difference between current time and most recent time a token has entered place $p$
\item [$\Delta_{max}(\mathcal{L})$] Maximum observed trace duration in an event log $\mathcal{L}$
\item [$E$ ] event instance vector
\item [$\varepsilon$] \textit{split miner} filtering threshhold hyperparameter
\item [$\eta$] \textit{split miner} parallelism threshhold hyperparameter
\item [$\mathcal{F}$] set of all \textit{arcs} of a PN
\item [$f_p(\tau)$] decay function of place $p$
\item [$F(\tau)$] decay function response vector
\item [$fn$] false negative
\item [$fp$] false positive
\item [$fpr$] false positive rate
\item [$g$] \textit{case} or \textit{trace} 
\item [$\mathcal{G}$] finite set of all possible \textit{traces}
\item [$\gamma(g)$] function returning the number of event instances of a trace $g$
\item [$l$] denotes a neural network layer in form of a matrix
\item [$\mathcal{L}$] event log which is a set of traces
\item [$|\mathcal{L}|$] number of traces of an event log $\mathcal{L}$
\item [$|\mathcal{L}_{i}|$] number of event instances  of the $i$th trace of an event log $\mathcal{L}$
\item [$\mathcal{L}_{i,j}$] $j$th event  instance  in the $i$th trace of an event log $\mathcal{L}$
\item [$M$] vector representing the \textit{marking} of a PN
\item [$M^{final}$] final marking
\item [$M^{init}$] initial marking
\item [$M_i$] $i$th element of marking $M$
\item [$M(\tau)$] vector representing the \textit{marking} of a PN at time $\tau$
\item [$\mathcal{M}$] set of all \textit{markings}
\item [$mean(\cdot)$] arithmetic mean function
\item [$\mathcal{N}$] set of all possible \textit{event instances}
\item [$\nu_p(g)$] number of tokens a place $p$ produces when replaying a trace $g$
\item [$p$] place
\item [$\mathcal{P}$]  set of all \textit{places}
\item [$|\mathcal{P}|$] cardinality of set of places
\item [$PN$] mathematical definition of a labeled PN
\item [$\phi_j(l_k)$] weighted input of neuron $j$ from previous layer $l_k$
\item [$\pi$] function which maps a transition to either  a single observable event or to the non-observable event
\item [$R(\tau)$] attribute value counting vector from time $0$ to $\tau$
\item [$\mathcal{R}$] set of all $\alpha_p$ of a PN
\item [$|\mathcal{R}|$] cardinality of the set $\mathcal{R}$
\item [$\rho_j(l_k)$] activation function based on input layer $l_k$
\item [$S(\tau)$] timed state sample at time $\tau$
\item [$\mathcal{S}$] set of timed state samples
\item [$\sigma(p)$] function returning number of tokens of a place $p$
\item [$t$] transition
\item [$tn$] true negative
\item [$tp$] true positive
\item [$tpr$] true positive rate
\item [$\mathcal{T}$] set of all \textit{transitions}
\item [$\tau$] time
\item [$\tau_p$] most recent time that a token entered place $p$
\item [$\theta_j(l_k)$] output function of a neuron based on input layer $l_k$
\item [$\upsilon_d(E)$] function returning value of \textit{attribute} $d$ of event  instance $E$
\item [$w_{i,j}$] weight of direct connection between two neurons $i$ and $j$
\item [$\bullet x$] set of input nodes of a node $x$
\item [$x \bullet$] set of output nodes of a node $x$
\item [$\xi$] cost function
\item [$\mathcal{Z}$] set of all non-negative integers\\

\item [$i, j, k, n, x, y$] indices, integers, and variables used in different contexts
\end{description}

\bibliography{sample.bib}{}
\bibliographystyle{IEEEtran}
\end{document}